\def\year{2019}\relax
\documentclass[letterpaper]{article} 
\usepackage{aaai19}  
\usepackage{times}  
\usepackage{helvet} 
\usepackage{courier}  
\usepackage[hyphens]{url}  
\usepackage{graphicx} 
\def\UrlFont{\rm}  
\usepackage{graphicx}  
\usepackage{varioref}
\usepackage{cleveref}
\usepackage[acronym]{glossaries}

\title{MuMMER: Socially Intelligent Human-Robot Interaction in Public Spaces}

\usepackage{xspace}
\newcommand{\gla}{\textsuperscript{\rm 1}\xspace}
\newcommand{\hwu}{\textsuperscript{\rm 2}\xspace}
\newcommand{\idiap}{\textsuperscript{\rm 3}\xspace}
\newcommand{\laas}{\textsuperscript{\rm 4}\xspace}
\newcommand{\sbre}{\textsuperscript{\rm 5}\xspace}
\newcommand{\vtt}{\textsuperscript{\rm 6}\xspace}

\author{Mary Ellen Foster,\gla 
Bart Craenen,\gla 
Amol Deshmukh,\gla
\\\Large\bf
Oliver Lemon,\hwu 
Emanuele Bastianelli,\hwu
Christian Dondrup,\hwu
Ioannis Papaioannou,\hwu
Andrea Vanzo,\hwu 
\\\Large\bf
Jean-Marc Odobez,\idiap
Olivier Can{\'e}vet,\idiap
Yuanzhouhan Cao,\idiap
Weipeng He,\idiap
\\\Large\bf
Angel Mart{\'i}nez-Gonz{\'a}lez,\idiap
Petr Motlicek,\idiap
R{\'e}my Siegfried,\idiap
\\\Large\bf
Rachid Alami,\laas
Kathleen Belhassein,\laas
Guilhem Buisan,\laas
Aur{\'e}lie Clodic,\laas
Amandine Mayima,\laas
\\\Large\bf
Yoan Sallami,\laas
Guillaume Sarthou,\laas
Phani-Teja Singamaneni,\laas
Jules Waldhart,\laas
\\\Large\bf
Alexandre Mazel,\sbre
Maxime Caniot,\sbre
\\\Large\bf
Marketta Niemel\"a,\vtt
P\"aivi Heikkil\"a,\vtt
Hanna Lammi,\vtt
Antti Tammela\vtt
\\
\gla University of Glasgow, Glasgow, UK 
\hspace{2ex} 
\hwu Heriot-Watt University, Edinburgh, UK  \\
\idiap Idiap Research Institute, Martigny, Switzerland 
\hspace{2ex}
\laas LAAS-CNRS, Toulouse, France \\
\sbre SoftBank Robotics Europe, Paris, France
\hspace{2ex}
\vtt VTT Technical Research Centre of Finland, Tampere, Finland\\
MaryEllen.Foster@glasgow.ac.uk
}

\nocopyright

\begin{document}

\maketitle

\begin{abstract}
In the EU-funded MuMMER project, we have developed a social robot designed to interact naturally and flexibly with users in public spaces such as a shopping mall. We present the latest
version of the robot system developed during the project. This system encompasses audio-visual sensing, social signal processing, conversational interaction, perspective taking, geometric reasoning, and motion planning. It successfully combines all these components in an overarching framework using the Robot Operating System (ROS) and has been deployed to a shopping mall in Finland interacting with customers. In this paper, we describe the system components, their interplay, and the resulting robot behaviours and scenarios provided at the shopping mall.
\end{abstract}

\maketitle
\section{Introduction}

In the EU-funded MuMMER project (\url{http://mummer-project.eu/}), we have developed a socially intelligent interactive robot designed to interact with the general public in open spaces, using SoftBank 
Robotics' Pepper humanoid robot 
as the primary platform
\cite{10.1007/978-3-319-47437-3_74}.
The MuMMER system provides an entertaining and engaging experience to enrich a human-robot interaction.
Crucially, our robot  exhibits
behaviour that is \textit{socially appropriate} and \textit{engaging} by combining
speech-based conversational interaction with non-verbal communication, and motion planning. 
To support this behaviour, we have developed and integrated new methods
from audiovisual scene processing, social-signal processing, conversational
AI, perspective taking, and geometric reasoning.

\begin{figure}
    \centering
   \includegraphics[width=.75\columnwidth]{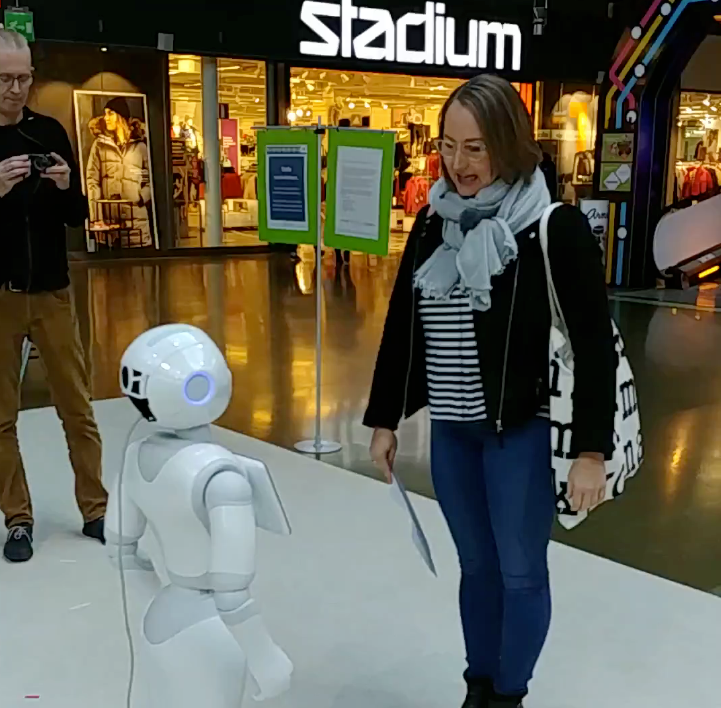}
    \caption{The MuMMER robot system interacting with a customer in the Ideapark shopping mall, September 2018.}
    \label{fig:pepper-ideapark}
\end{figure}

The primary MuMMER deployment location is Ideapark, a large shopping mall in
Lemp\"a\"al\"a, Finland. 
The MuMMER robot system has been taken to the shopping mall several times for short-term 
co-design activities with the mall customers and retailers~\cite{Heikkilae2018,vttlaas}; the full robot system has been deployed for short periods in 
the mall in September 2018 (Figure~\ref{fig:pepper-ideapark}), May 2019, and June 2019, and 
has been installed for a long-term, three-month deployment as of September 2019. 

The demo system supports a range of behaviours
covering a variety of functional and entertainment tasks that are
appropriate for a shopping-mall setting, including guidance to various locations within the mall, 
small-talk, and playing quiz games with customers.
The activities during the deployment have  
included a number of data collection studies with real users: recording of customer 
interaction with the robot in guidance situations, sound localisation and automatic speech recognition in the 
noisy mall environment, and tests for AI-based conversation and localisation and 
navigation based on a partial 3D model of the mall and a complete semantic model.

In the remainder of this paper, we outline the technical contributions in each of the
main MuMMER component areas: audiovisual sensing, social signal processing, conversational interaction,  human-aware robot motion planning, knowledge representation and decision. At the end, we describe the details of the deployed robot system.

\section{Audio-visual sensing}\label{sec:audio-visual-sensing}

For MuMMER, the main task of audio-visual perception is sensing people in
general -- that is, maintaining a representation of the persons around
the robot, with a dedicated attention to people susceptible of
interacting with it, or those who are (or have been) interacting
with it. This requires several audio-visual algorithms
to detect, track, re-identify people, and detect their
non-verbal behaviors and activities, and also predict their
position/behaviours even when they are not seen. At the same time, the
representation of people needs to be defined and shared with other
modules which are responsible for inferring other knowledge about
people (for instance, to define a person's goal in the interaction).

\begin{figure}
\includegraphics[width=\columnwidth]{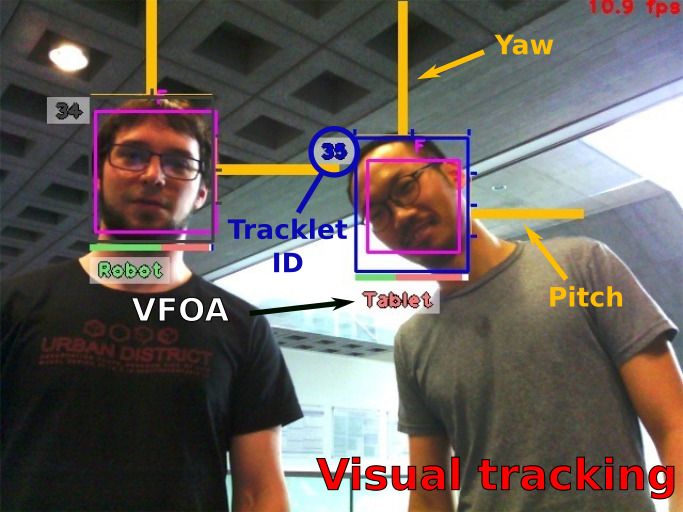}
\includegraphics[width=\columnwidth]{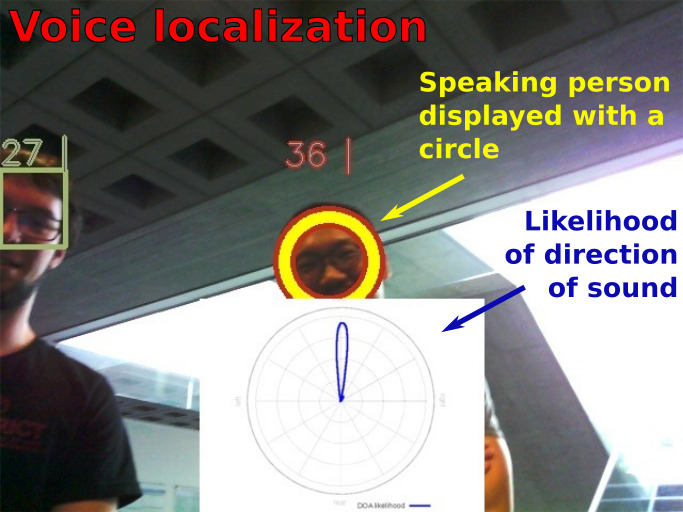}
\caption{The perception systems tracks and re-identifies 
people leaving the field, and extracts other features: speaker turns, 
head pose, visual focus of attention and nods.}
\label{fig:visual-tracking}
\end{figure}

For visual tracking, we first detect the person with the
convolutional pose machines (CPM)~\cite{cao2017realtime} which
provide accurate locations of the body joints (nose, eyes, shoulders,
etc.). The output of the algorithm is almost perfect when people are
in the foreground of the image and up to 3 of 5 meters, depending on
the resolution. This is our use-case definition of an entertainment robot in a
shopping mall. On top of the CPM, we use
OpenHeadPose\footnote{\url{https://gitlab.idiap.ch/software/openheadpose}}
\cite{cao2018leveraging}, which makes use of the heap maps of the CPM
to estimate the head pose of the person. Then, we perform head pose
tracking~\cite{Khalidov_Idiap-RR-02-2017} to maintain a consistent
identity across adjacent frames
(Figure~\ref{fig:visual-tracking}). The head pose tracker is represented by a
particle filter mainly based on color and face cues. As faces are
tracked, we store OpenFace features~\cite{amos2016openface} which are
computed on an aligned face. When a new tracklet is created, the
OpenFace features of the new tracklet are compared with the features
previously accumulated, and the new tracklet is re-assigned the
identity of the one which had the more votes.

For sound localization , we use a multi-task neural network (NN)
which jointly performs speech/non-speech detection and sound source
localisation~\cite{he2018deep,he2018joint} applied on top of the 4-channel microphone
array (embedded on the robot). The NN uses as an input a 4-channel audio transformed into a frequency domain, and it outputs the likelihood values for the two tasks. 
Thanks to a semi-automated and synthetic data collection procedure taking advantage of the robotic platform as well as the use of a weak supervision learning approach, it is possible to quickly collect data to learn the models for a new sensor~\cite{He:ICASSP:2019}.
The fusion between the visual and audio parts is done by assigning the detected
speech to the person who is standing in the given direction.

Finally, although a close range (up to 1.2m) gaze sensing module is available and can be applied for one selected person using a self-calibrated approach \cite{Siegfried:ICMI:2017}, as a compromise between computation and robustness, we instead compute the visual focus of attention of each person
based on the head pose~\cite{sheikhi2015combining}. 
The algorithm can reliably estimate the object the person is looking at (either the robot,
the other persons, the targets, the shops, or the tablet embedded at the robot)
which is a preliminary step to identify the addressee, and is also used in the context 
of perspective taking to determine whether the human has looked in the direction where
the robot pointed~\cite{Sallami-iros2019}.

\section{Social signal processing}

\begin{figure}[!ht]
    \centering
    \includegraphics[width=.45\textwidth]{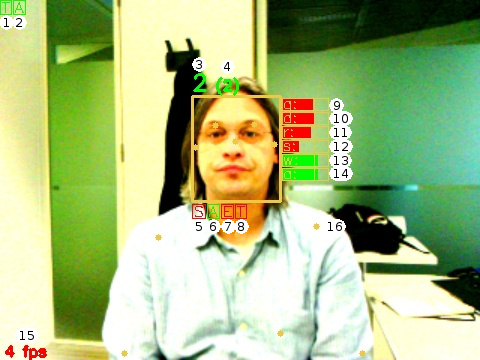}
    \caption{Social state estimator visualiser output, displaying all relevant information for fine-tuning.}
    \label{fig:fusion_vis}
\end{figure}

\begin{figure}
\centering
\begin{tabular}{ccc}
\includegraphics[width=0.12\textwidth]{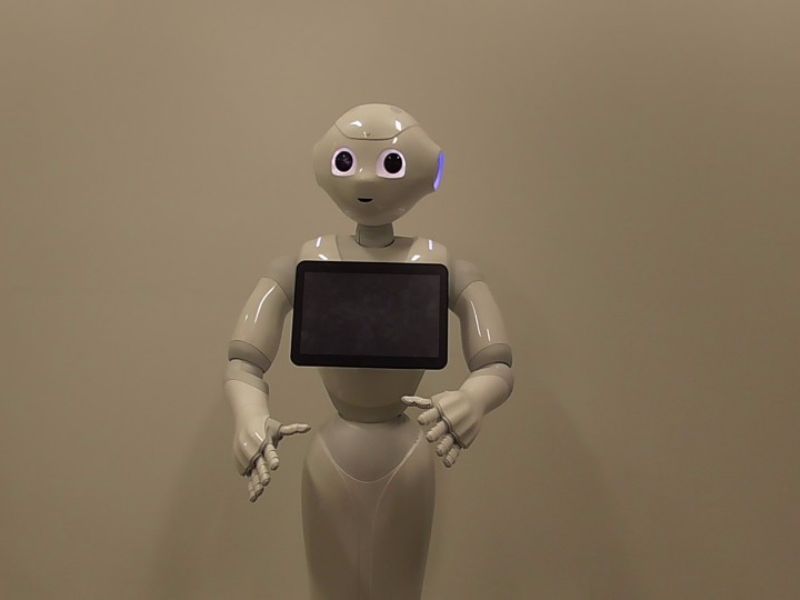} &
\includegraphics[width=0.12\textwidth]{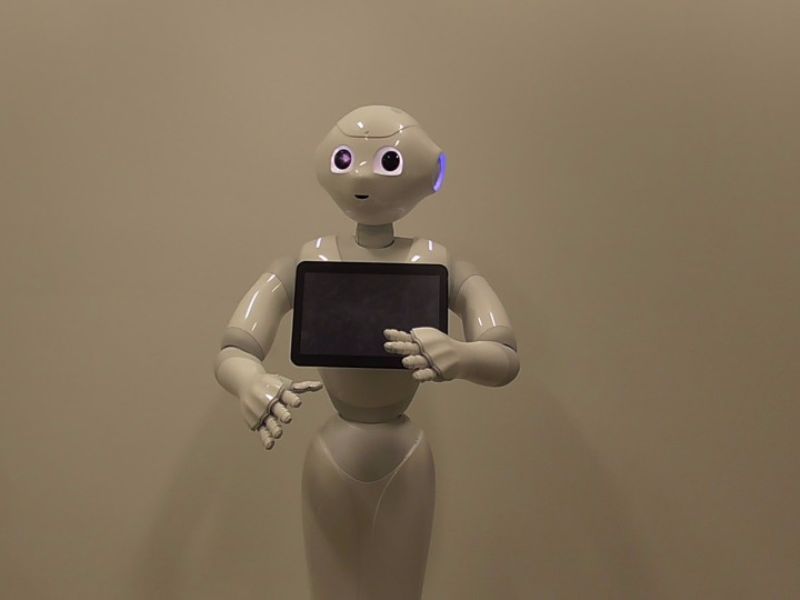} &
\includegraphics[width=0.12\textwidth]{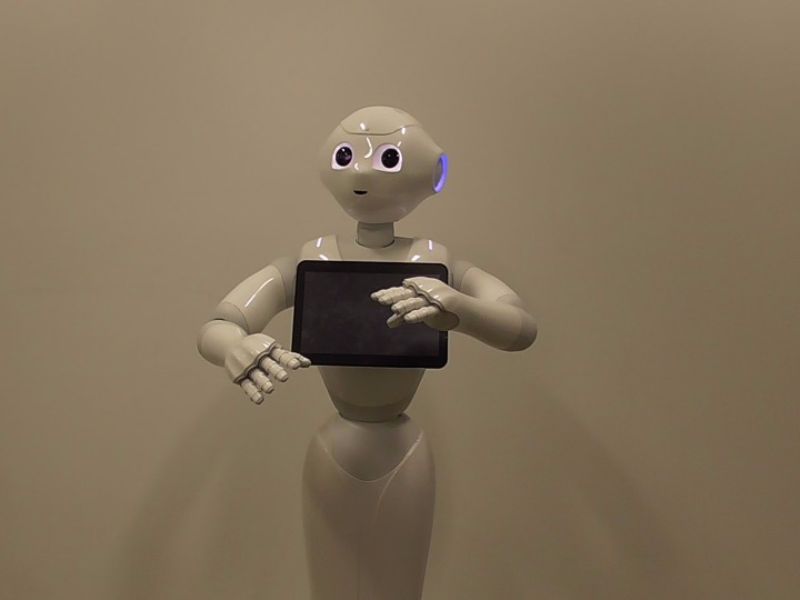} \\
\includegraphics[width=0.12\textwidth]{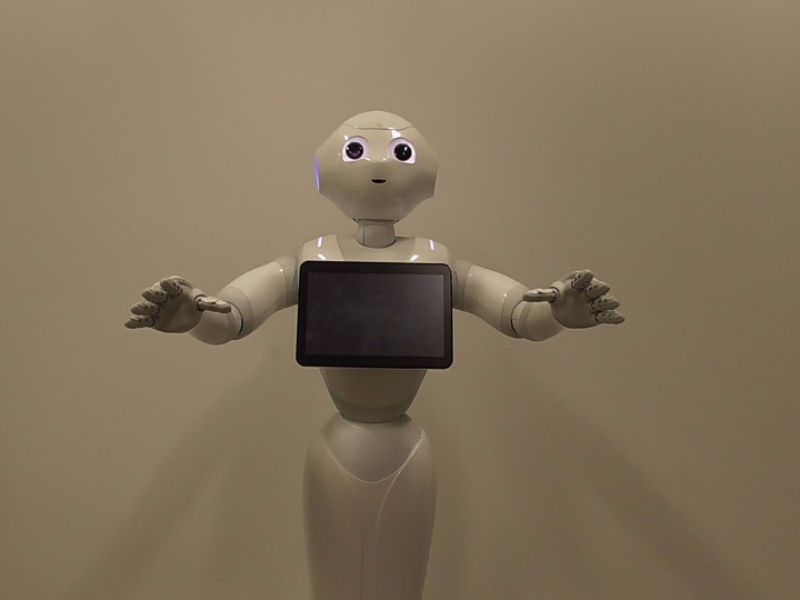} &
\includegraphics[width=0.12\textwidth]{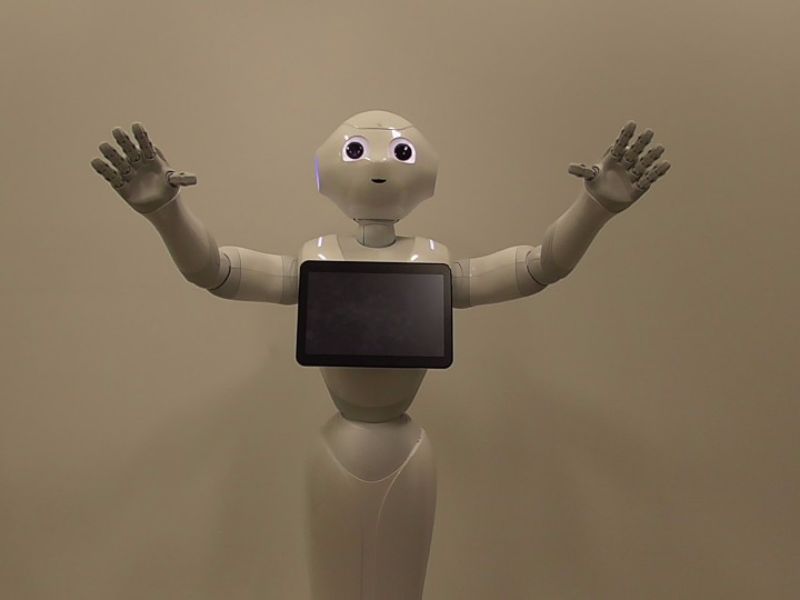} &
\includegraphics[width=0.12\textwidth]{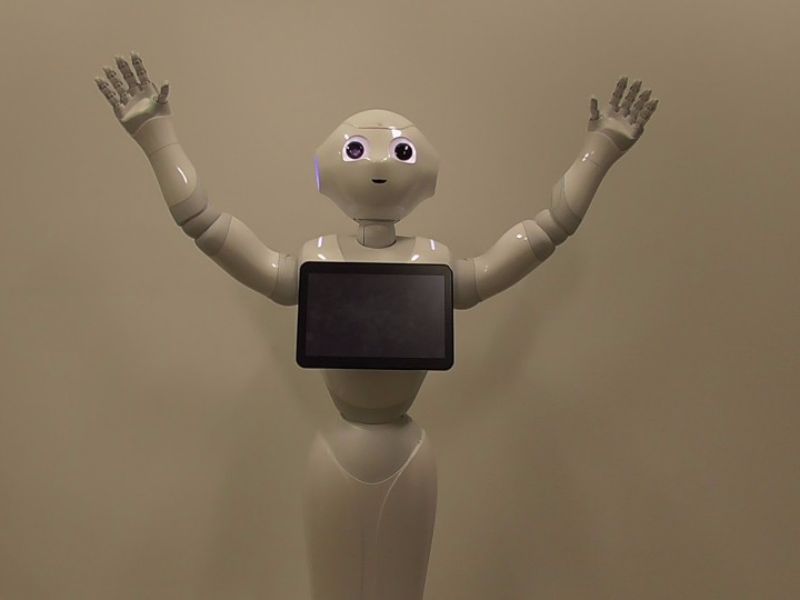} \\
\includegraphics[width=0.12\textwidth]{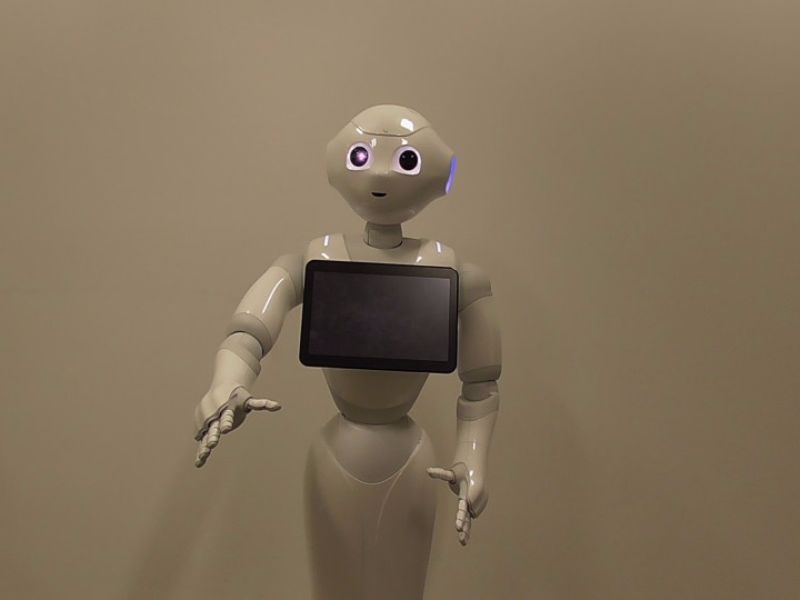} &
\includegraphics[width=0.12\textwidth]{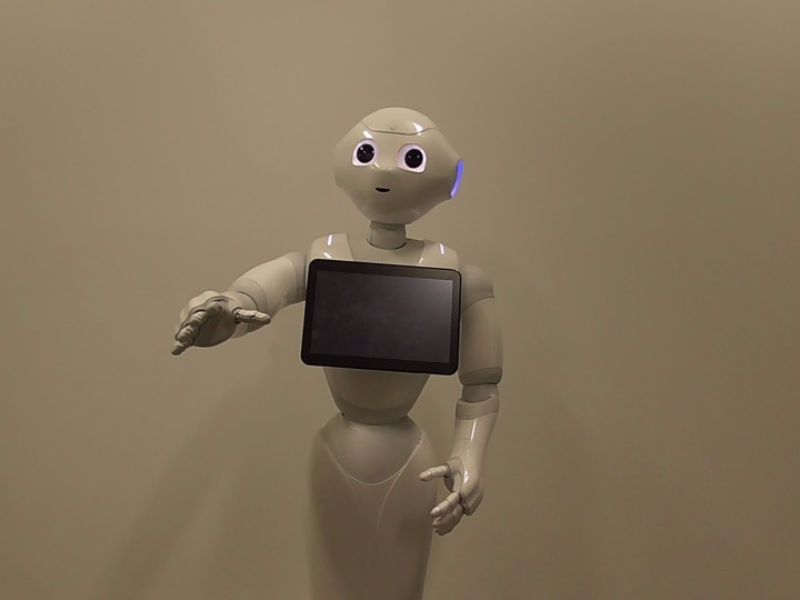} &
\includegraphics[width=0.12\textwidth]{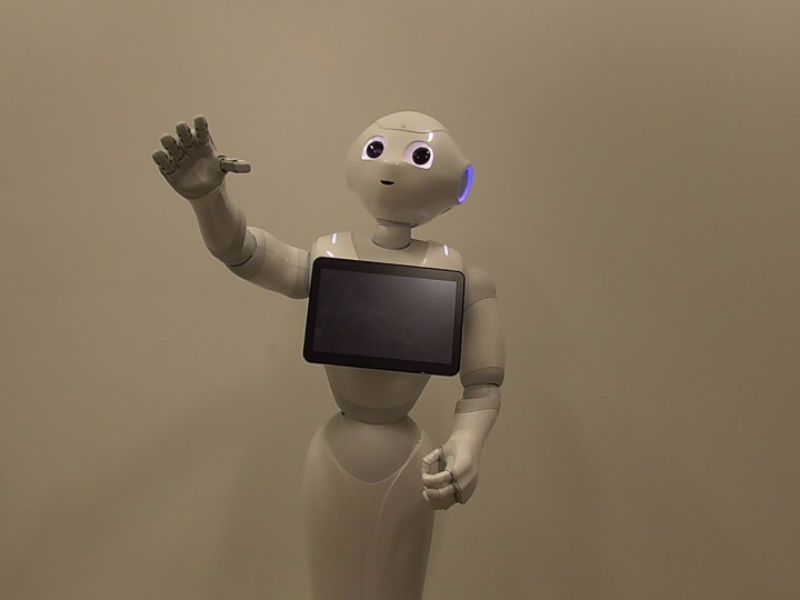} \\
\includegraphics[width=0.12\textwidth]{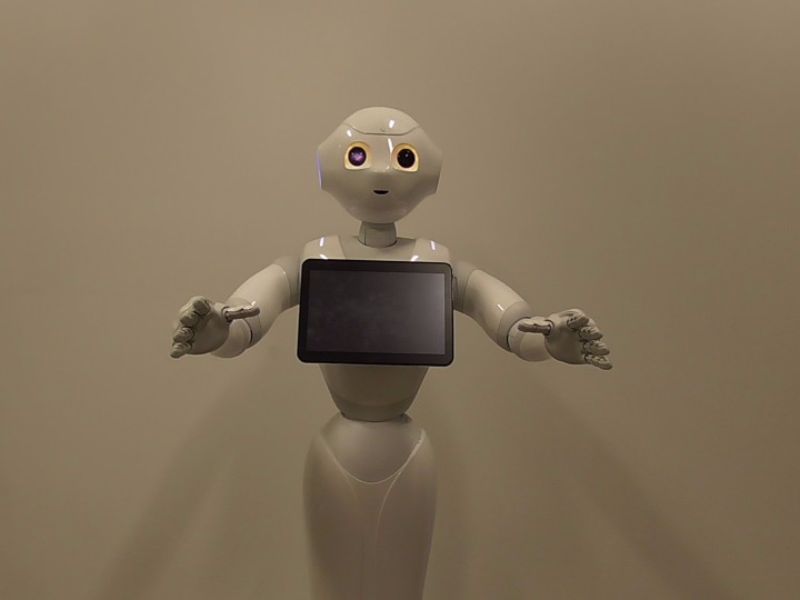} &
\includegraphics[width=0.12\textwidth]{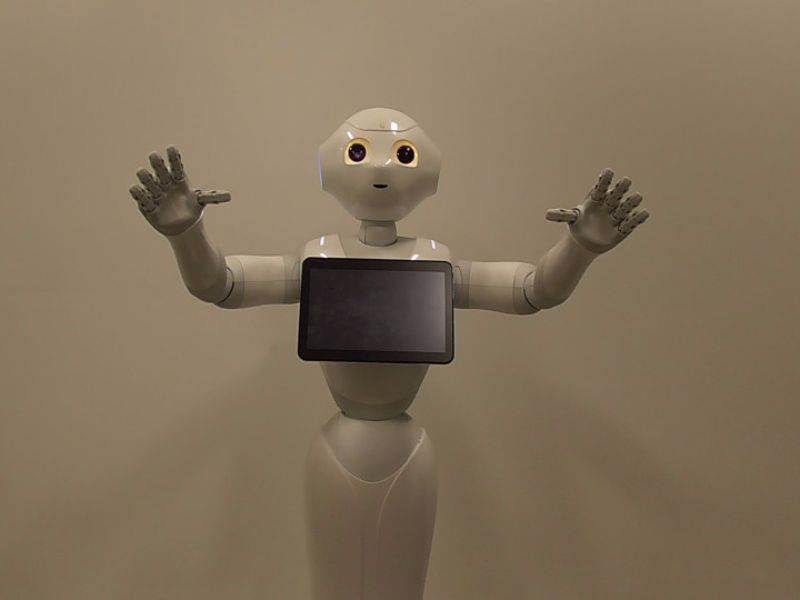} &
\includegraphics[width=0.12\textwidth]{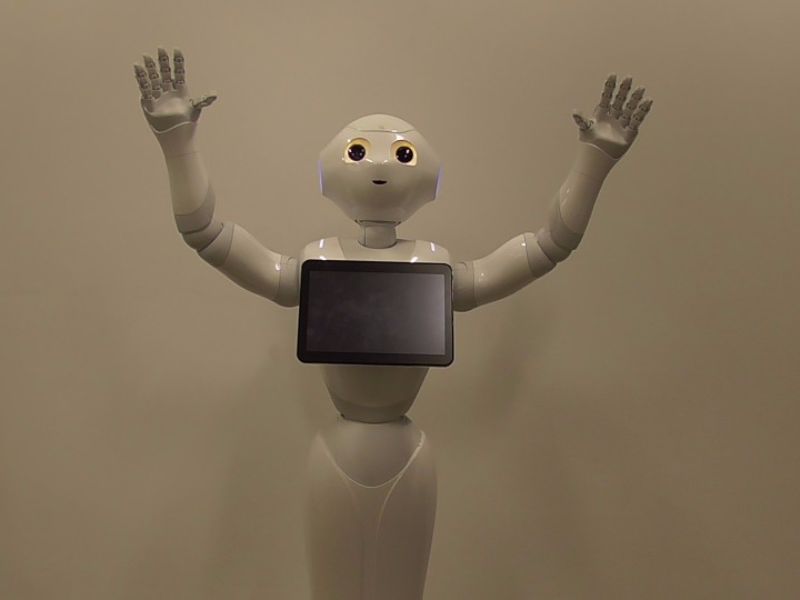} \\
\end{tabular}
\caption{Gesture variations. Each row shows the same gesture using different parameters.}\label{tab:gestures}
\end{figure}
For Social Signal Processing, we focus on two primary tasks: fusing the provided audio-visual sensing data for social state estimation, and synthesising appropriate social signals for the robot to use when communicating with users. 
While detecting, tracking, (re)identifying users, as well as detecting their primary non-verbal behaviours and activities provide the basic signals, the multi-modal fusion of these signals allows for a more accurate and deeper understanding of the underlying social state, including gaining personality impressions from the user. 
The estimated social state is then made available to inform planning of the robot's subsequent actions; who and how to converse with the users of the robot; and, how the robot is to move and behave (gestures) in the presence of the user(s). 

\subsection{Social state estimation}
On the fusion side, the main function of the social state estimator is to determine which user the robot should initiate interaction with.
We used the underlying assumption that the robot should initiate interaction with the user perceived to be the most willing to interact; which we took to be the user paying the most attention to the robot.
We assume that the user paying the most attention to the robot is the user that is looking most directly at the robot, and who is most closely situated near the robot.

To this end, the social state estimator aggregates audio-visual sensing data about the head pose of users, whether the users are looking at the robot and/or the screen on the robot, and the distance between the users' head and the robot. 
The head pose data of the users are used to calculate the (Euclidean) distance between the head pose of the users and three centroids derived from clustering/classifying previously recorded lab and deployment data.
This distance is then normalised to a value between zero and one, and used as a probability.
The distances between the users and the robot are used as a penalty, and normalised between zero and one as a probability in such a way that users further away from the robot are penalised more than users closer to the robot.

This results in four probabilities, two taken directly from the audio-visual sensing data, and two derived from it.
These four probabilities are then fused into one \emph{attention} probability by calculating their (weighted) average.
Choosing which user the robot should interact with is done by comparing the attention probability against a configurable minimum attention probability threshold, and then selecting the user with the highest attention probability.

To prevent immediately re-initiating an interaction with a user that the robot has just interacted with, the social state estimator also monitors the actions of the planning and dialogue components.
The social state estimator then maintains a list of the users that the robot is, and has been, interacting with, and applies a penalty to their attention probability while they are interacting and for a short time afterwards.

The social state estimator is fully configurable by a set of parameters, with the initial parameter settings determined from extensive recorded lab data.
The parameters were further fine-tuned during deployment to provide the most accurate and applicable social state estimates.
To facilitate fine-tuning the parameters of social state estimator, a visualiser is provided to display all relevant features (Figure~\ref{fig:fusion_vis}).

\subsection{Social signal generation}

For the synthesis side, a repertoire of non-verbal social signals, including gestures and sounds, has been developed for the robot, available to be used in conjunction with moving and interacting with the users. The non-verbal behaviour of an embodied agent is at least as communicative as its verbal behaviour \cite{Vinciarelli_Pantic_Bourlard_2009}, and in 
a noisy environment such as a mall it may even be more important, so understanding
and controlling the robot's non-verbal signals is crucial.
Examples of some robot gesture variants are shown in Figure~\ref{tab:gestures}. 

In a series of perception experiments, we have examined how manipulating gesture parameters affect users' subjective responses to the robot as well as their perception of the robot's personality. 
These studies have found several clear relationships: for example, manipulating the amplitude and speed significantly affected users' perception of the Extraversion and Neuroticism of the robot, while the attributed personality also affected users' subjective reactions to the robot \cite{craenen_ea_ROMAN_2018}. In addition,
it was found that while the majority of users preferred a robot that they perceived
to have a similar personality to their own, a significant minority preferred a robot whose personality was perceived to be
different than their own \cite{8525672}. 

We are currently integrating a finer-grained method of gesture control based on sentiment \cite{deshmukh-etal-roman2019}, as well as a set of affectively generated artificial sounds \cite{Hastie:2016:SEA:2993148.2993169}, with the goal of further enhancing the robot's expressiveness.

\section{Conversational interaction}
\label{sec:convinteraction}
\begin{figure}[t]
\includegraphics[width=\columnwidth]{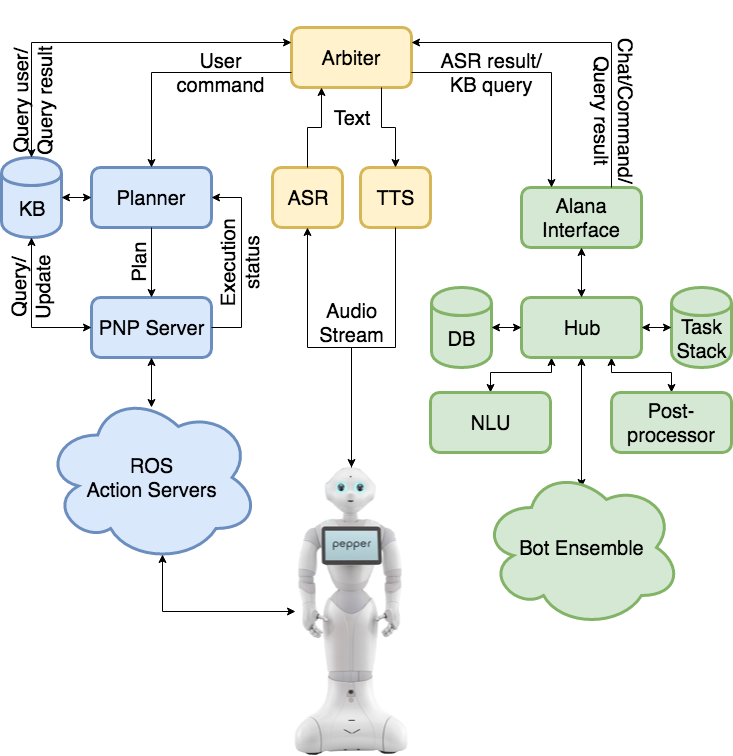}
\caption{Architecture of the Dialogue system. The blue parts on the left represent the task management and execution system, the green parts on the right represent Alana as the dialogue system. The Bot Ensemble contains social chat bots and the \textit{task bot} which is able to trigger tasks and handle communication between the task and the user. The yellow middle part is the task specific dialogue management system (Arbiter), Text-To-Speech (TTS) and Automatic Speech Recognition (ASR).}
\label{fig:alana}
\end{figure}

The MuMMER system focuses on enabling an agent to combine a task-based dialogue system with chat-style open-domain social interaction, to fulfil the required tasks while at the same time being natural, entertaining, and engaging to interact with. The presented work is based on the ``Alana'' conversational framework, a finalist of the Amazon Alexa Challenge in both 2017~\cite{papaioannou2017alana} and 2018~\cite{curryalana}. Alana was initially developed for the Amazon Echo as an open-domain social chatbot. For the needs of this project, Alana acts as the core module for every dialogue interaction with the user from every other module. This means that whenever a module requires to either verbally notify the user or get the user's feedback, Alana will handle this task. In this way the conversation throughout the interaction will be more contextually relevant, and easier to maintain. Since the robot needs to engage in social dialogue as well as to complete tasks, Alana was enriched with so-called \textit{task bots} to conversationally execute and monitor behaviours on a physical agent (Figure~\ref{fig:alana})~\cite{papaioannou2018human}. 

In order to enable the functionality described above, a new Natural Language Understanding (NLU) module,  HERMIT NLU, has been implemented and integrated into the Alana system, which is able to deal with social chit-chat but also extract the necessary information from commands to start tasks. HERMIT NLU~\cite{vanzo:2019b} is thus used to decide if the \textit{task bot} is triggered and to extract the required parameters for tasks,  such as the name of the shop someone is looking for. While standard chatbots mostly rely on NLU that works on shallow semantic representations (e.g., intents + slots), 
task-based applications require richer characterisations. In line with~\cite{dinarelli2009annotating}, we promote the idea that the user's intent can be represented through the combination of existing theories, capturing different dimensions of the overall problem, namely \textit{Dialogue Acts} and \textit{Frame Semantics}.
Existing approaches to NLU for dialogue systems are based on formal languages designed around the targeted domain.
However, it has been widely demonstrated that the generalisation capability of statistical-based approaches is more robust towards lexical and domain variability~\cite{bastianelli:2016a}.
We thus use a deep learning architecture based on a hierarchy of self-attention mechanisms and BiLSTM encoders followed by CRF tagging layers to perform multi-task learning over the aforementioned semantic dimensions~\cite{rastogi2018multi}.
The system effectively learns how to predict Dialogue Acts, Frames, and Frame Elements in a sequence labelling scheme, starting from a corpus of annotated sentences which we are currently developing.

After a task has been identified, executing it on a robot usually includes physical actions that require a finite amount of time to complete and are not instantaneous such as dialogue actions. While the robot is executing such an action, the user might want to continue the conversation, or give new instructions. In order to be able to support such a multi-threaded dialogue management of interleaving tasks with general chitchat and other tasks, we build on the ideas presented in~\cite{Lemon:2002:Multi-tasking}. To this end, the execution system introduced in~\cite{dondrup2017introducing} has been extended to use so-called \textit{recipes} that define dialogue and physical actions to execute in order to achieve the given goal~\cite{papaioannou2018human}. The execution framework described therein has been redesigned to support multi-threaded execution and an arbitration process has been put in place to manage the currently running tasks on the execution side and in the Alana system. This lets tasks be started, stopped, and paused at any time, with appropriate feedback to the user. If a task has been suspended by another action, it will be resumed after the new action finishes and any open questions will be  re-raised to prompt the user.

\section{Route guidance supervision}

One of the core tasks for the MuMMER robot in the mall is the guiding of users to specific locations in the mall, by pointing at places and explaining the route to the wanted location. The task is triggered when a human asks for a location. A supervision system, based on Jason~\cite{Bordini:2007:PMS:1197104}, a BDI agent-oriented framework handles the execution through Jason reactive plans. 
Throughout the task, the robot supervises the execution and, depending what goes wrong, the robot has multiple possible responses. For example, it is able to handle nominal scenarios of route guidance while being able to take into account contingencies such as the human lack of visibility of the direction, his/her ability or not to take the stairs, his/her understanding of the message, etc. Finally, if at some point, the human is not perceived during a certain time, the robot ends the task, assuming that the human has left. 

\subsection{Route computing and route verbalization} \label{ssec:route_computing}
The entire description of the route, from the search for the best route to get to the final destination to the verbalization of this route, is based on the SSR (Semantic Spatial Representation)~\cite{sarthou-splu19}. This representation is used to describe the topology of an indoor environment as well as semantic information (type of stores or items sold by stores) in a single ontology. This ontology is managed by Ontologenius~\cite{sarthou-roman19}, a lightweight open-source ROS-compatible package which stores semantic knowledge, reasons with it, and shares that information to all the other system components.

\subsection{Geometric reasoning} \label{ssec:geo_reasoning}
\begin{figure}
\includegraphics[width=\columnwidth]{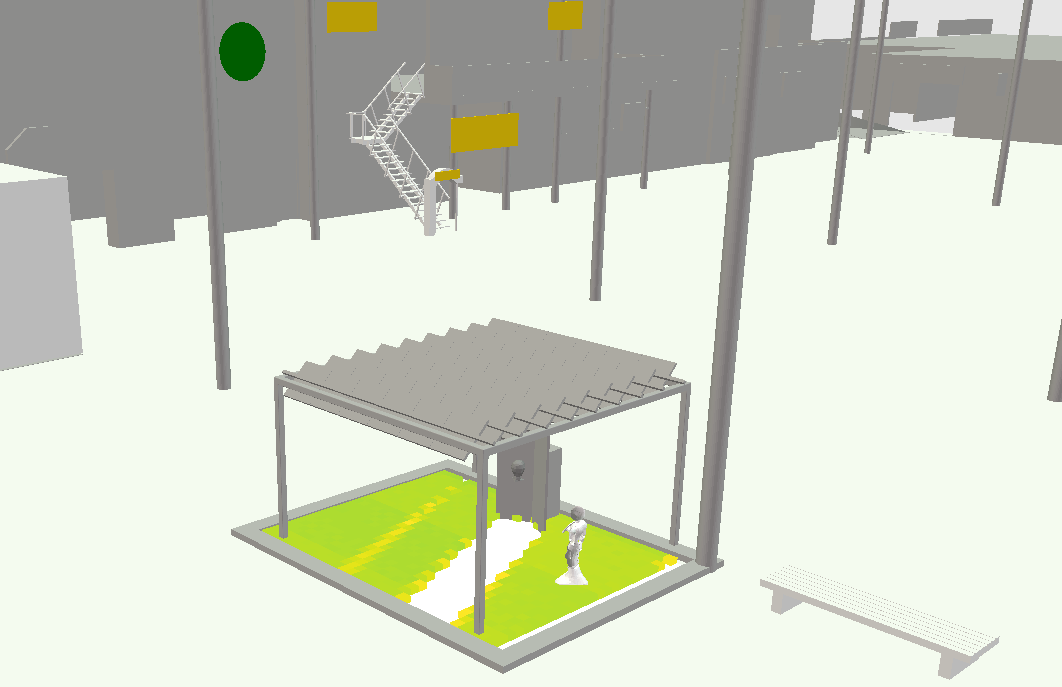}
\caption{Visualization of the visibility grids of a landmark on the 3D model of the central square of the Ideapark shopping center.}
\label{fig:ideapark-model}
\end{figure}

Geometric reasoning uses Underworlds~\cite{lemaignan-2018,Sallami-iros2019}, a lightweight framework for \emph{cascading spatio-temporal situation assessment} in robotics. It represents the environment as real-time distributed data structures, containing scene graph (for representation of 3D geometries). 
Underworlds supports \emph{cascading} representations: the environment is viewed as a set of \emph{worlds} that can each have different spatial granularities, and may inherit from each other. It also provides a set of high-level client libraries and tools to introspect and manipulate the environment models. 
Based on a 3D model of the mall (Figure~\ref{fig:ideapark-model}), it maintains what the robot knows about the scene as well as alternative world states. These states represent the estimation of the human's beliefs about the scene
It also provides the symbolic relations among entities with stamped predicates (e.g. [$isInsideArea(person,area)$] or [$isSpeakingTo(X,Y)$] when $X$ speaks and looks at $Y$ (given by perception)). 

\subsection{Motion planning} \label{ssec:motion_planning}

The navigation of the robot is implemented using the ROS navigation stack, with navfn as the global planner and a Timed Elastic Band (TEB)~\cite{rosmann2017integrated} planner as the local planner. For MuMMER, the local planner was modified in order to accommodate humans into planning inspired from~\cite{khambhaita_isrr17}, resulting a new planner called and this new planner called Social TEB (S-TEB).
This algorithm is able to plan and execute trajectories while ensuring satisfaction of robot kinematics constraints, avoiding static and moving non-human obstacles and planning navigation solutions respecting social constraints with humans perceived. The planner ensures the safety of humans by re-planning a local plan at each control loop. 

\subsection{SVP planner} \label{ssec:svp}
Although the target robot location in the mall is in a large square, several elements of the environment can block the visibility of important landmarks for the proper understanding of the route to take. The purpose of the SVP (Shared Visual Perspective) planner~\cite{waldhart:hal-02283904} is therefore to try to find a position where the human will have to go in order to observe an element of the environment such as a passage, a staircase or a store. To do this, a visibility grid is computed for each possible landmark, as shown on figure~\ref{fig:ideapark-model}. Having determine a good position for the human, the planer also allows to determine the good position for the robot so as to have a human-robot-landmark conformation allowing both to point the landmark and to look at the human.

\section{Deployment}

A long-term deployment (three months from September 2019) will allow the study of the customer behaviours
around a helpful and entertaining robot over an extended period of time. This section gives details on how 
the hardware and software that are being employed in the final deployment, as well as the 
scenarios that are supported. 

\subsection{Setup}

The fully integrated MuMMER system consists of several hardware
components to allow the computation to be performed on the appropriate
platforms.

The robot we are using is an updated, custom version of the Pepper platform,
which is equipped with an Intel D435 camera and a NVIDIA Jetson TX2
in addition to the traditional sensors that are found on the previous
versions of the robot. We
use the Robot Operating System (ROS) to enable the communication
between the processing nodes. All the streams (audio, video,
robot states) are sent to a remote laptop which performs all the
computation. The laptop has a NVIDIA RTX 2080 graphics card (for the
deep learning part) and 12 CPU cores. The perception
algorithms 
process the Intel images at a
resolution of $320 \times 180$ for the detection and tracking parts,
and at a resolution of $640 \times 360$ for the re-identification part, 
which enables fast tracking and a good re-identification quality
with OpenFace. The 4 microphone streams are processed at a frequency of
$16000$~Hz, and the full perception system delivers the output
at 10~fps.

To transcribe the user's speech signal we use the Google Automatic Speech Recognition (ASR) API\footnote{\url{https://cloud.google.com/speech-to-text/}} which receives an enhanced audio signal from a delay and sum beamformer based on the location of the speaking person determined by the audio-visual sensing. A dedicated ROS node streams the audio to the ASR which in real-time returns an incrementally updated string transcribing the utterance. Using silence to mark the end of speech, this transcription is enhanced using the context of the sentence to provide for a more coherent result. Finally, the text output of the Google ASR is sent to the Alana framework to perform the dialogue task, through the arbitration module as explained 
above.

The system is deployed in two languages, English and Finnish, though due to the vast linguistic differences between the two languages, the two versions have been kept separated, and the whole interaction can either be in one or the other. Due to the complexity of the NLU module in the Finnish version of the system, the user's utterance is being translated into English using Google Translate API\footnote{\url{https://cloud.google.com/translate/}}. The result of this translation is sent to the Alana conversational framework and goes through the NLU pipeline described in detail in \cite{curryalana}. In the English version of the system, Alana then returns the reply to be verbalised. Due to the relatively poor performance of Google translate when it comes to translating English into Finnish (as remarked upon by our Finnish partners), the Finnish version of Alana has a much reduced set of bots in its ensemble (see Figure~\ref{fig:alana}). These bots mainly return answers based on templates that have been translated into Finnish beforehand.

\subsection{Scenarios deployed}

As a proof of concept, a real-time autonomous system has been 
built to integrate all the components described in the sections
above. The following types of interactions can be triggered by the user:

\begin{description}
\item[Chat] The staple of the interaction is social dialogue~\cite{curryalana}. During all other modes of interaction, the user can always default to simply chat to the robot irrespective of whether it is currently executing a goal/task (e.g.\ the user requires guidance to a specific shop) or not. For example the user might approach the robot and start discussing various topics. At specific points throughout this conversation, the system might explain its capabilities to the user in order to recover from a conversational stalemate or to simply make them aware of the fact that it can also be helpful in finding your way around the mall (see below).
\item[Quiz] In this scenario, multiple choice questions are
asked by the robot, and the human replies by stating the
number of the answer they think is correct.
\item[Route Description (Dialogue only)] When the human asks how to get to a specific shop, the robot gives him/her the route description. In this most ``simple'' form, the system uses only verbal interaction. This means that especially the route description is merely presented as a string of synthesised text.
\item[Route Guidance (Dialogue + Pointing)] In this version, the robot guides the human to specific locations in the mall, by pointing at places and explaining the route to the wanted location. To do so, the robot first computes positions so that the human will be able to see what the robot is pointing for him/her. Then, the robot navigates to its position (this part is optional), expecting that the human will join it once it stopped and checks the human's visibility. Then, the robot explains to the human how to reach the destination. According to a human-human guidance study~\cite{belhassein:hal-01719730}, the robot points, first at the location direction and then points at the access point (a corridor, stairs or an escalator) to go through to reach the location. While pointing, the robot verbalizes the route. Finally, the robot checks that the human knows how to reach the goal and leaves open the possibility to repeat if needed. All along the task, the robot supervises the task and adapts accordingly. 
\end{description}

All these modes of interaction can be interleaved at the user's discretion. This means, for example, that during the quiz the user could revert to social dialogue. If they do so, the system might occasionally try to bring them back to the quiz by re-raising the last question. The same holds true if the person chooses to abandon a route guidance task before it was finished.

\section{Conclusions}

The MuMMER project has built a fully autonomous entertainment
robot to perform HRI scenarios in a shopping mall, in which the 
main goal is to have entertainment interaction (quiz, chat), as well
as route guidance. The system is real-time, by leveraging the
heavy deep learning computation on a remote laptop, the 
ASR on the Google platform, and the Alana conversational AI system on a
remote server. This system enables a natural
interaction with the participants; it has been tested and was tested in 
real conditions for several short sessions, and as of September 2019 is
fully deployed for a three-month long-term user study.

Further work for large scale deployment could include
some software optimizations
to run more components on the robot itself, and to reduce the
lag which sometime exists between the human speech and the robot
reply.

\section{Acknowledgements}
This research has been partially funded by the European Union's Horizon 2020 research and innovation program under grant agreement no. 688147 (MuMMER, \url{http://mummer-project.eu/}).

\clearpage

\bibliographystyle{aaai}
\bibliography{references}

\end{document}